\def\year{2020}\relax
\documentclass[letterpaper]{article} 
\usepackage{aaai20}  
\usepackage{times}  
\usepackage{helvet} 
\usepackage{courier}  
\usepackage[hyphens]{url}  
\usepackage{graphicx} 
\usepackage{graphicx}  
\usepackage{algorithm}
\usepackage{algorithmic} 
\usepackage{amsmath}
\usepackage{amsfonts}
\usepackage{multirow}
\usepackage{makecell}
\usepackage{bm}

\title{MixPoet: Diverse Poetry Generation via Learning\\ Controllable Mixed Latent Space}

\author{
Xiaoyuan Yi,\textsuperscript{\rm 1} 
Ruoyu Li,\textsuperscript{\rm 3} 
Cheng Yang,\textsuperscript{\rm 2} 
Wenhao Li,\textsuperscript{\rm 1} 
Maosong Sun\textsuperscript{\rm 1}\thanks{\quad Corresponding author: M.Sun (sms@mail.tsinghua.edu.cn)}\\
\textsuperscript{\rm 1}Department of Computer Science and Technology, Tsinghua University\\ 
Institute for Artificial Intelligence, Tsinghua University\\ 
State Key Lab on Intelligent Technology and Systems, Tsinghua University\\ 
\textsuperscript{\rm 2}Beijing University of Posts and Telecommunications\\
\textsuperscript{\rm 3}6ESTATES PTE LTD, Singapore\\
\{yi-xy16, liwh16\}@mails.tsinghua.edu.cn, sms@tsinghua.edu.cn\\
yangcheng@bupt.edu.cn, ruoyuli1995@gmail.com 
}

\begin{document}

\maketitle

\begin{abstract}
As an essential step towards computer creativity, automatic poetry generation has gained increasing attention these years. Though recent neural models make prominent progress in some criteria of poetry quality, generated poems still suffer from the problem of poor diversity. Related literature researches show that different factors, such as life experience, historical background, etc., would influence composition styles of poets, which considerably contributes to the high diversity of human-authored poetry. Inspired by this, we propose \emph{MixPoet}, a novel model that absorbs multiple factors to create various styles and promote diversity. Based on a semi-supervised variational autoencoder, our model disentangles the latent space into some subspaces, with each conditioned on one influence factor by adversarial training. In this way, the model learns a controllable latent variable to capture and mix generalized factor-related properties. Different factor mixtures lead to diverse styles and hence further differentiate generated poems from each other. Experiment results on Chinese poetry demonstrate that MixPoet improves both diversity and quality against three state-of-the-art models.
\end{abstract}
\section{Introduction}
\label{sec_intro}
\begin{figure}
\centering
\includegraphics[scale=0.38]{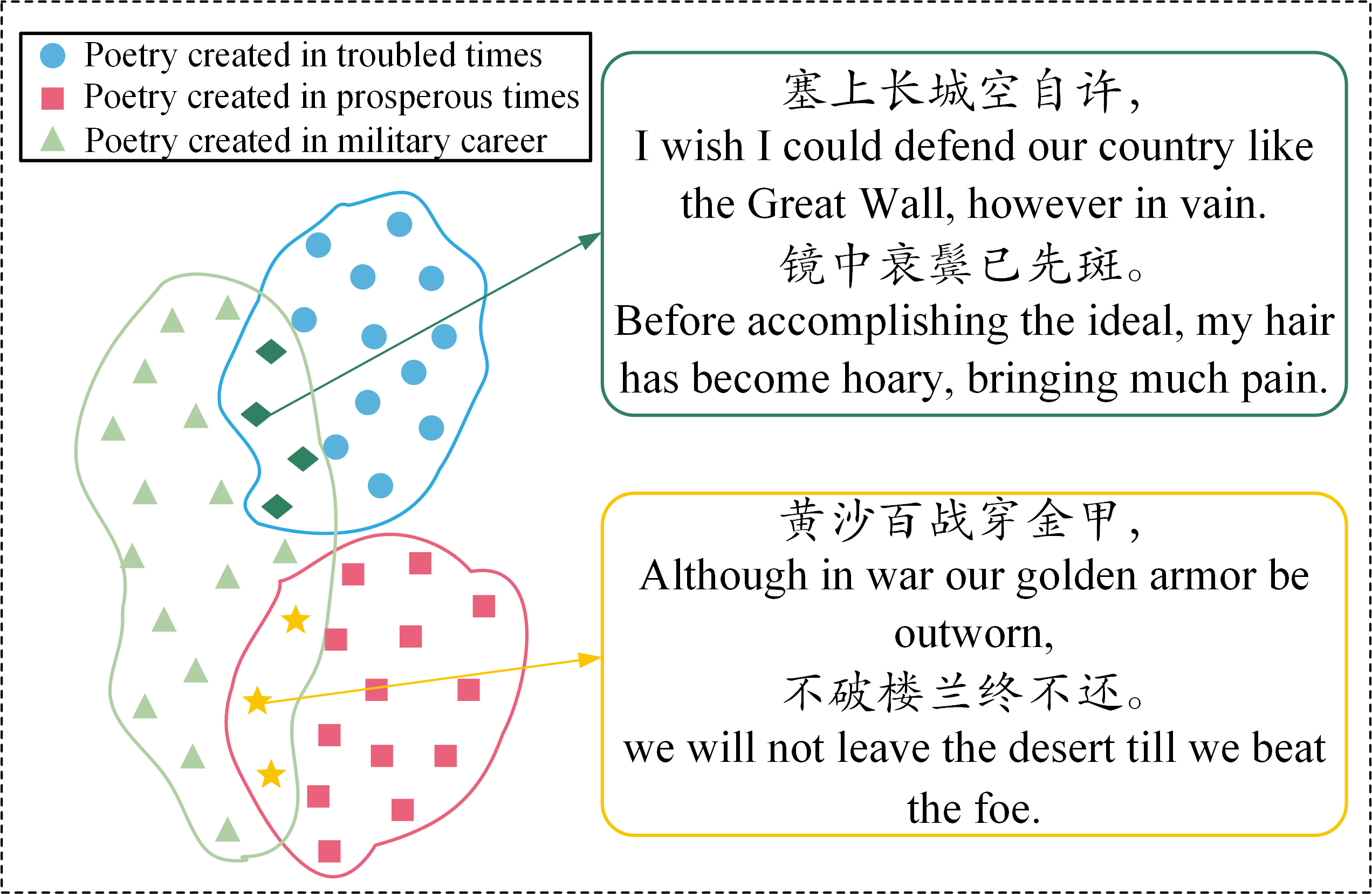}
\caption{Left: an illustration of the poetry spaces on different influence factors. Right: two human-authored sentences (each with two lines) absorbing two factors.}
\label{fig_case1}
\end{figure}
Poetry is one of the most valuable cultural heritages for human beings. Characterized by its elegant expressions, colorful contents and diverse styles, this literary genre appeals to people across different ages and nationalities. Automatic poetry generation has attracted growing attention in the past several years because of its considerable research value in exploring computer creativity and building humanizing AI, which could also benefit the construction of intelligent assistants for entertainment and education.

Recent models mainly make efforts and achieve significant progress in improving some primary criteria of poetry quality, such as context coherence~\cite{Yanpolish:16} and topic relevance~\cite{Marjantopical:16,Liadv:18}. However, beyond these criteria, generated poems still suffer from the problem of \emph{poor diversity}.

Intuitively, the fundamental requirements of diversity in poetry generation could be two-fold: (1) poems generated with different topic words should be distinguishable from each other (\emph{inter-topic} diversity) and (2) with the same topic word, the model should be able to generate distinct poems (\emph{intra-topic} diversity). Nevertheless, most existing models fail to meet such requirements since they tend to remember some common patterns in the corpus and produce repetitive and generic contents, even with different topic words as input~\cite{Zhangmemory:17,Yimrl:18}.

To address this problem, we must figure out what contributes to diversity. Related literature theories demonstrate that different factors would influence writing manners of human poets, such as their life experience~\cite{dilthey1985poetry}, historical background~\cite{owen1990poetry}, school of literary, etc. These factors lead to differences in thoughts, feelings, and expressions in poetry composition, which underlie the diverse styles of poets and make human-authored poems highly distinguishable, as observed in \cite{Zhangmemory:17}. Figure~\ref{fig_case1} gives an example: under the same topic (the war), the poem created by a poet who lived in a powerful and prosperous dynasty tends to express strong confidence and aspiration; by contrast, the other created by a poet living in troubled times shows the sorrow and worry of being invaded.

Inspired by this, we propose a novel model, \emph{MixPoet}, which absorbs different influence factors to improve the diversity of generated poems. To exploit the underlying properties of factors, we resort to semi-supervised Variational AutoEncoder (VAE)~\cite{Kingma:14c}. We don't assume the independence of the latent variable and influence factors because poetry style is tightly coupled with semantics~\cite{Wellerstyle:67}. Instead, our model disentangles the latent space into some subspaces and makes each conditioned on one factor (together with the keyword) by adversarial training, to capture and mix generalized factor-related semantics. In the training phase, our model can predict factors of unlabelled poems and thus be trained in a semi-supervised manner. In the testing phase, by specifying different values for each factor, we can create various mixtures of factor properties that bring distinctive new styles for generated poems.

With the same given keyword, by manually varying the mixture, one can create distinct poems that simultaneously express properties of multiple factors, achieving intra-topic diversity. With different keywords as input, our model can automatically infer an appropriate factor mixture for each keyword and thus generate more distinguishable poems, improving inter-topic diversity.

In summary, our contributions are as follows:
\begin{itemize}
\item To the best of our knowledge, we are the first effort at generating poems that mix the properties of different factors for the sake of better diversity.
\item We innovatively propose a semi-supervised MixPoet model to disentangle the latent space into different factor-conditioned subspaces by adversarial training. 
\item We experiment on Chinese poetry. Automatic and human evaluation results show that our model can controllably mix different factors and improve both the diversity and quality of generated poems, against three current state-of-the-art models.
\end{itemize}
\section{Related Work}
\label{sec_relawork}
As an important chapter of automatic natural language generation, poetry generation has interested researchers for decades. After the early attempt of template-based models~\cite{Gervas:01}, systems based on statistical machine learning methods, such as genetic algorithms~\cite{Manurung:03} and statistical machine translation approaches~\cite{He:12}, make the first breakthrough and generate barely satisfactory poems.

The past several years have witnessed the rapid progress of neural networks, which also show notable advantages in poetry generation. Existing works mainly target at improving some primary criteria of poetry quality. At first, the Recurrent Neural Network (RNN) is used to generate fluent poems~\cite{Zhangrnn:14,Hopkinsverse:17}. After that, pursuing better context coherence, the Polish model~\cite{Yanpolish:16} embellishes a generated poem several times. To enhance topic relevance, the Hafez system~\cite{Marjantopical:16} extracts more related keywords to bring more abundant topic information; the working memory model~\cite{Yimemory:18} leverages an internal memory to store and access multiple topic words.

Despite the significant improvement on these criteria, models mentioned above fail to meet a higher requirement, the diversity. To handle this problem, the MRL model~\cite{Yimrl:18} uses reinforcement learning to encourage high-TF-IDF words, which improves inter-topic diversity. The USPG model~\cite{Yang:18a} generates stylistic poetry by maximizing the mutual information between styles and poems, which promotes intra-topic diversity. Since USPG is trained in an unsupervised manner, the learned styles are indistinguishable and uninterpretable.

VAE has recently proven to be effective for generating various types of text~\cite{Zhaovaedialog:17,Zhangvaetrans:16}. Related to our work, Yang et al.~\shortcite{Yangvaepoetry:18} use VAE to learn a context-conditioned latent variable for poetry generation. Hu et al.~\shortcite{hutoward:17} suppose the independence of latent space and attributes to generate single sentences but without constraints on semantics. Li et al.~\shortcite{Liadv:18} use adversarial training to match generated poems and given titles to strengthen topic relevance.

Our motivation and method considerably differ from these works. For better diversity, we apply adversarial training to the latent space (instead of explicit poems) and disentangle it into factor-conditioned (neither factor-independent nor context-conditioned) subspaces to involve various styles and generate diverse poems under the control of both required topic and factors. Besides, our model is semi-supervised and can be trained well with a fraction of labelled data.
\section{Model}
\label{sec_model}
Before detailing the proposed MixPoet, we first formalize our task. Define $x$ as a poem with $n$ lines $x_1,x_2,\dots,x_n$, each line with $l_i$ words as $x_i \! = \! x_{i,1},x_{i,2},\dots,x_{i,l_i}$, and $w$ as a keyword representing the main topic. Suppose there are $m$ factors, $y_1,\dots,y_m$. Since influence factors are quite complicated concepts, to simplify the problem, we discretize each factor $y_i$ into $k_i$ classes. By specifying different classes (values) for each factor, we can create $\prod_{i=1}^m k_i$ factor mixtures, with each leading to a new distinctive style. As poems with manually annotated factor labels are rare, we also utilize unlabelled data and define $p_l(x,w,y_1,y_2,\dots,y_m)$ and $p_u(x,w)$ as the empirical distributions over labelled and unlabelled datasets respectively. Our goal is to generate poems which are relevant to $w$ on topic and concurrently accord with the mixed factors on style.
\subsection{Basic Generator}
\label{subsec_basic}
\begin{figure}
\centering
\includegraphics[scale=0.29]{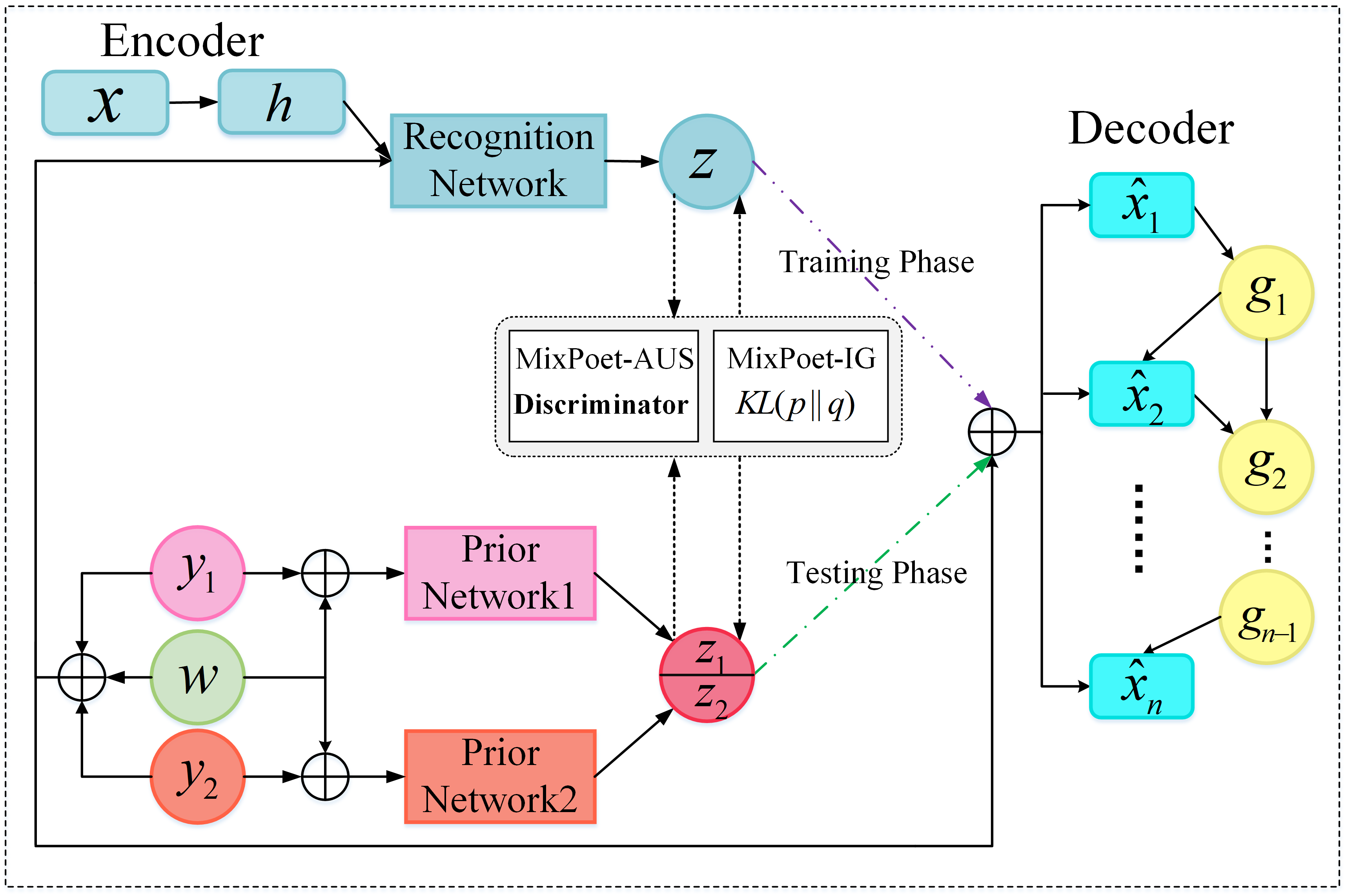}
\caption{A graphical illustration of MixPoet. The latent variable is sampled from the posterior distribution for training and from the prior one for testing.}
\label{fig_model}
\end{figure}
We first present a basic generator, one of our baselines, which is also a part of MixPoet. We adopt an effective structure similar to that in \cite{Yanpolish:16,Yimrl:18}.

Define $s_{i,j}$ as the corresponding GRU~\cite{Cho:14} decoder hidden state. Then the probability distribution of each $x_{i,j}$ to be generated is computed as:
\begin{align}
& s_{i,j} \! = \! GRU(s_{i,j-1}, [e(x_{i,j-1}); g_{i-1}]), \\
& s_{i,0} = f(e(w), o_i), \\
& p(x_{i,j}|x_{i,<j},x_{<i},w) \!=\! softmax(f(s_{i,j})),
\end{align}
where $[;]$ means concatenation; $e(\cdot)$ represents the embedding; $x_{<i}$ is the abbreviation of $x_1,\dots,x_{i-1}$ (similar to $x_{i,<j}$); $o_i$ is a special length embedding~\cite{Yimemory:18} to control the length of each line; $f$ is a non-linear layer.

$g_{i-1}$ is a context vector to record so-far generated content in a poem and provide global information for the generator, which is used to keep context coherence and computed as:
\begin{align}
& a_{i,t} = f([s_{i,t};\dots;s_{i,t+d-1}]), \\
& g_i = f(g_{i-1}, \sum_{t} a_{i,t}), g_0=\bm{0},
\end{align}
where $\bm{0}$ is a zero vector and $d$ is the window size.
\subsection{Semi-Supervised Conditional VAE}
\label{subsec_semivae}
We introduce the semi-supervised framework of our model, which is adopted in our previous work~\cite{Chenemotion:19}. We first give the formalization based on a single factor $y$ for brevity and will incorporate more factors later.

Aiming at learning the conditional joint distribution $p(x,y|w)$, we can involve a latent variable $z$ and have $p(x,y|w) \! = \! \int p(x,y,z|w) dz$. Since style is coupled with semantics as mentioned in Sec.~\ref{sec_intro}, rather than suppose the independence of $z$ and $y$, we decompose $p(x,y,z|w)$ as $p(x,y,z|w)=p(y|w)p(z|w,y)p(x|z,w,y)$. Such decomposition indicates how a poem is generated: if the user doesn't provide any label, the model predicts an appropriate factor class by the keyword, then draws a sample of $z$ according to the required topic ($w$) and factor ($y$), and finally generates a poem ($x$). During this process we could manipulate both the topic and style of generated poems by separately specifying the keyword and factor class.

Then for labelled data, we can derive the lower bound:
\begin{equation}
\begin{split}
& \mathbb{E}_{q_{\phi}(z|x,w,y)}[\log\,p_{\psi}(x|z,w,y)] \\
& -KL[q_{\phi}(z|x,w,y)||p_{\theta}(z|w,y)] \\
& + \log\,p_{\omega}(y|w) = -\mathcal{L}(x,y,w) \leq \log p(x,y|w), 
\end{split}
\label{eq1}
\end{equation}
where we approximate the true prior distribution $p(z|w,y)$ and posterior distribution $q(z|x,w,y)$ with a prior network $p_{\theta}(z|w,y)$ and a recognition network $q_{\phi}(z|x,w,y)$ respectively. $\theta$ and $\phi$ are corresponding parameter sets.

By optimizing Eq.(\ref{eq1}), we reconstruct the poem $x$, and minimize the KL divergence of the posterior and prior distributions. Besides, we also incorporate a classifier $p_{\omega}(y|w)$ to predict appropriate factor classes when the user doesn't provide any label. $\omega$ represents the parameters of classifiers.

Since the labelled data is too limited to train the model well, as \cite{Kingma:14c}, to utilize unlabelled poems, we treat the unobserved $y$ as another latent variable. In a similar vein, we can derive and maximize:
\begin{equation}
\begin{split}
& \mathbb{E}_{q_{\omega}(y|x,w)}[ \! -  \! \mathcal{L}(x,y,w)] \! + \! \mathcal{H}(q_{\omega}(y|x,w)) \\
& \! = \! -\mathcal{U}(x,w) \leq \log p(x|w),
\end{split}
\label{eq2}
\end{equation}
where another classifier $q_{\omega}(y|x,w)$ is trained to infer classes for unlabelled poems with Gumbel-softmax \cite{Janggumbel:17} during the training process.

Ultimately, the total semi-supervised loss is:
\begin{equation}
\begin{split}
\mathcal{L} &= \mathbb{E}_{p_l(x,w,y)}[\mathcal{L}(x,y,w) - \alpha * \log\,q_{\omega}(y|x,w)] \\
& + \beta * \mathbb{E}_{p_u(x,w)}[\mathcal{U}(x,w)],
\end{split}
\label{eq3}
\end{equation}
where we also add the classification loss to the first term to train the classifier $q_{\omega}(y|x,w)$ utilizing both supervised and unsupervised signals. $\alpha$ and $\beta$ are hyper-parameters.

Figure~\ref{fig_model} diagrams our model. In detail, we take the whole poem $x$ as a long sequence and feed it into a bidirectional GRU. Then we concatenate the last forward and backward hidden states to form $h$, the vector representation of $x$. The classifiers are implemented with Multi-Layer Perceptron (MLP): $p_{\omega}(y|w)=softmax(MLP(e(w)))$ and $q_{\omega}(y|x,w)\! = \!softmax(MLP(e(w),h))$. We refer to $p_{\psi}(x|z,w,y)$ as the decoder (parameterized by $\psi$), which is just the basic generator introduced in Sec.~\ref{subsec_basic}, except that we set the initial decoder state as $s_{i,0}=f(e(w),o_i,z,e(y))$ to involve the latent variable and the factor.
\subsection{Latent Space Mixture}
The formulas above only focus on a single factor. To incorporate $m$ factors, we can assume that the latent space can be disentangled into $m$ subspaces $z\!=\![z_1;\cdots;z_m]$. Without loss of generality, we give the formulation when $m$=$2$. By further assuming the independence of influence factors and the conditional independence of these subspaces, we have $p(z|w,y)\!=\!p(z_1|w,y_1)p(z_2|w,y_2)$. Accordingly, we need to replace the classifiers in Sec.~\ref{subsec_semivae} with $p_{\omega}(y_1|w)$, $p_{\omega}(y_2|w)$, $q_{\omega}(y_1|x,w)$ and $q_{\omega}(y_2|x,w)$ to predict $y_1$ and $y_2$ respectively. This disentanglement indicates that we can independently draw $z_1$ and $z_2$ from corresponding factor-conditioned subspaces to form the whole latent variable. That is, we get a latent space which mixes the properties of different factors. We design two methods to learn such mixed latent space.
\subsubsection{Mixture for Isotropic Gaussian Space}
We call the first method \textbf{MixPoet-IG} since we assume the latent variable follows the isotropic Gaussian distribution as previous related works~\cite{Kingma:14c,Yangvaepoetry:18} usually do.

Then we can rewrite the KL divergence in Eq.(\ref{eq1}) as $KL[q_{\phi}(z_1|x,w,y_1)||p_{\theta}(z_1|w,y_1)] + KL[q_{\phi}(z_2|x,w,y_2)||p_{\theta}(z_2|w,y_2)]$. Since $z_1$ and $z_2$ also follow the isotropic Gaussian distribution, we implement the recognition networks and the prior networks with MLP, for example, $p_{\theta}(z_1|w,y_1) \! \sim \! \mathcal{N}(\mu_1, \sigma_1^2\bm{I})$ where $[\mu_1;\log \sigma_1^2] = MLP(e(w),e(y_1))$. Then we can analytically minimize these two KL terms, draw samples of latent variables with the reparametrization trick~\cite{Kingma:14} and train the whole model with Eq.~(\ref{eq3}).
\begin{algorithm}[t]  
\begin{algorithmic}[1] 
\FOR {number of iterations}
\STATE Sample labelled batch $\{x,w,y_1,y_2\}$;
\STATE Sample unlabelled batch $\{x,w\}$ and sample corresponding predicted labels $y_1 \sim q_{\omega}(y_1|x,w) $, $y_2 \sim q_{\omega}(y_2|x,w) $;
\STATE Sample the posterior latent variable $z$ and the prior $z_1,z_2$ with Eq.(\ref{eq9});
\STATE Train the four classifiers ($\omega$), recognition network ($\phi$) and decoder ($\psi$) in Eq.(\ref{eq3});
\STATE Train the discriminator ($\upsilon$) with Eq.(\ref{eq11});
\STATE Adversarially train the recognition network ($\phi$) and prior networks ($\theta$) with Eq.(\ref{eq12});
\ENDFOR
\end{algorithmic}
\caption{Training Process of MixPoet-AUS} 
\label{alg1}
\end{algorithm}
\subsubsection{Adversarial Mixture for Universal Space}
The second method is called \textbf{MixPoet-AUS}. Despite the tractability of computation, the isotropic Gaussian distribution may fail to learn more complex representations as discussed in \cite{Dilokthanakul:17}. We want to only keep the independence of $z_1$ and $z_2$ but with the internal dimensions of each subspace entangled. By this means, the model can learn more generalized latent representations with enough capacity to hold the broad concepts of influence factors and meanwhile control them independently.

Therefore, we don't specify any concrete form of the latent space. Instead, we use a universal approximator~\cite{makhzani2015adversarial} and make the model learn arbitrary complex forms by itself. In detail, for a conditional distribution $q(z|c)$ with a condition $c$, we assume:
\begin{equation}
q(z|c,\eta) = \delta(z-MLP(c,\eta)),
\label{eq9}
\end{equation}
where $\eta$ is random noise and $\delta$ is the impulse function. By replacing $c$ with a certain condition (\textit{e.g.}, $w,y_1$) and sampling $\eta \sim \mathcal{N}(\bm{0}, \bm{1})$ we can get samples of required latent variables (\textit{e.g.}, $z_1$).

Then we use the density ratio loss \cite{rosca2017variational} to approximate the KL term as follows:
\begin{equation}
\begin{split}
& KL[q_{\phi}(z|x,w,y_1,y_2)||p_{\theta}(z_1|w,y_1)p_{\theta}(z_2|w,y_2)] \\ 
& =\mathbb{E}_{q_{\phi}(z|x,w,y_1,y_2)}[\log \frac{q_{\phi}(z|x,w,y_1,y_2)}{p_{\theta}(z_1|w,y_1)p_{\theta}(z_2|w,y_2)}] \\
& \approx \mathbb{E}_{q_{\theta}(z|x,w,y_1,y_2)}[\log \frac{\mathcal{C}_{\upsilon}(z,y_1,y_2)}{1-\mathcal{C}_{\upsilon}([z_1;z_2],y_1,y_2)}],
\end{split}
\end{equation}
where $\mathcal{C}_{\upsilon}$ is a latent discriminator (parameterized by $\upsilon$) which discriminates between latent values sampled from the posterior distribution and the ones independently sampled from the two factor-conditioned prior distributions.

As in \cite{mohamed2016learning,Zhaoarae:18}, we use adversarial training to minimize this ratio loss which alternately optimizes the discriminator by:
\begin{equation}
\resizebox{1.0\linewidth}{!}{$
    \displaystyle
\begin{aligned}
& \mathop{\max}_{\upsilon} \mathbb{E}_{p_{\theta}(z_1|w,y_1)p_{\theta}(z_2|w,y_2)}[\log \! ( 1  - \! \mathcal{C}_{\upsilon}([z_1 ; z_2],y_1,y_2))]  \\ 
& +\mathbb{E}_{q_{\phi}(z|x,w,y_1,y_2)}[\log \mathcal{C}_{\upsilon}(z,y_1,y_2)],
\end{aligned}
$}
\label{eq11}
\end{equation}
and trains the recognition and prior networks by:
\begin{equation}
\begin{split}
& \mathop{\max}_{\phi,\theta} \mathbb{E}_{q_{\theta}(z_1|w,y_1)q_{\theta}(z_2|w,y_2)}[\log \mathcal{C}_{\upsilon}([z_1;z_2],y_1,y_2)] \\
& - \mathbb{E}_{q_{\phi}(z|x,w,y_1,y_2)}[\log \mathcal{C}_{\upsilon}(z,y_1,y_2)].
\end{split}
\label{eq12}
\end{equation}
In this adversarial training, we consider the prior network as a `generator' and the latent values sampled from the recognition network as `real data' in the standard Generative Adversarial Networks~\cite{Goodfellowgan:14}. The complete training process is shown in Algorithm \ref{alg1}.

When the discriminator is successfully cheated ($\mathcal{C}_{\upsilon}(\cdot)\!\approx\!0.5$), the KL divergence can be minimized close to zeros. In this way, the model learns a sophisticated latent space and disentangles it into different factor-conditioned subspaces. In Sec.~\ref{sec_experiment}, we will show that compared to Mixpoet-IG, Mixpoet-AUS learns more distinguishable latent representations and achieves better diversity.
\begin{table}
\center
\small
\begin{tabular}{c|c|c|c|c|c}
\Xhline{1.2pt}
\# of & MC & CL & Others & UNK & Total  \\
\hline
PT & 799 & 608 & 675 & 9,052  &   11,134\\
TT & 1,481 & 977 & 1,122 & 8,993  & 12,573 \\
UNK & 8,547  & 9,543 & 7,654 & - & 25,744   \\
\hline
Total & 10,827 & 11,128 & 9,451 & 18,045  & 49,451   \\
\Xhline{1.2pt}
\end{tabular}
\caption{Statistics of CQCF. MC: military career, CL: countryside life, PT: prosperous times, TT: troubled times. `Others' means poems that don't belong to MC or CL. `UNK' means unknown. We detail the collection methodology of CQCF in the supplementary file.}
\label{table_dataset}
\end{table}
\subsection{Training}
For MixPoet-IG, to alleviate the vanishing latent variable problem in VAE training, besides the annealing trick~\cite{Bowman:16}, we also add a BOW loss~\cite{Zhaovaedialog:17} to Eq.(\ref{eq3}) to force $z$ to capture more global information. For MixPoet-AUS, since the discriminator is a crucial part for adversarial training, we adopt a powerful projection discriminator recently proposed in \cite{Miyatocgan:18} and apply the spectral normalization~\cite{Miyatospec:18} to the discriminator to stabilize the training process.
\section{Experiments}
\label{sec_experiment}
\begin{table}[htp]
\center
\small
\begin{tabular}{c|c|c|c}
\Xhline{1.2pt}
Models  & inter-JS $\downarrow$ & intra-JS $\downarrow$ & LMS $\uparrow$  \\
\Xhline{1.2pt}
fBasic & - & 9.15\% & 0.34 \\
Basic & 2.58\% & -  & 0.31  \\
CVAE & 2.34\% & 38.2\%  & 0.33  \\
USPG & 1.89\% & 5.01\% & 0.37 \\
MRL & \bf 1.28\% & - & 0.33  \\
\hline
MixPoet-IG & 1.55\% & 8.35\% & 0.37   \\
MixPoet-AUS & 1.39\% & \bf 3.73\% & \bf 0.39 \\
\hline
GT & 0.12\% & - & 0.68 \\
\Xhline{1.2pt}
\end{tabular}
\caption{Automatic evaluation results of diversity. inter-JS: inter-topic Jaccard similarity. intra-JS: intra-topic Jaccard similarity. For calculating inter-JS, USPG and MixPoet predict appropriate styles/mixtures in terms of keywords.}
\label{table:diversity}
\end{table}
\subsection{Data}
We mainly experiment on two typical factors: \emph{living experience} and \emph{historical background}. We discretize the first one into three classes: military career, countryside life and others; and the second one into two classes: prosperous times and troubled times. By mixing these factors, we can create six new styles. Then we build a labelled corpus called Chinese Quatrain Corpus with Factors (CQCF), which contains 49,451 poems, and each poem is labelled on at least one of the two factors. Statistics of CQCF are reported in Table \ref{table_dataset}. Besides, we also collect a Chinese Quatrain Corpus (CQC) as unlabelled data which comprises 117,392 poems. For CQC, we randomly select 4,500 poems for validation and testing, respectively, and the rest for training. For CQCF, we use 5\% for validation, 5\% for testing.

We use TextRank~\cite{Mihalcea:2004} to extract keywords from poems to build $<$keyword, poem$>$ pairs and $<$keyword, poem, labels$>$ triplets, as in \cite{Yimrl:18}.
\subsection{Setups}
\begin{figure}
\centering
\includegraphics[scale=0.63]{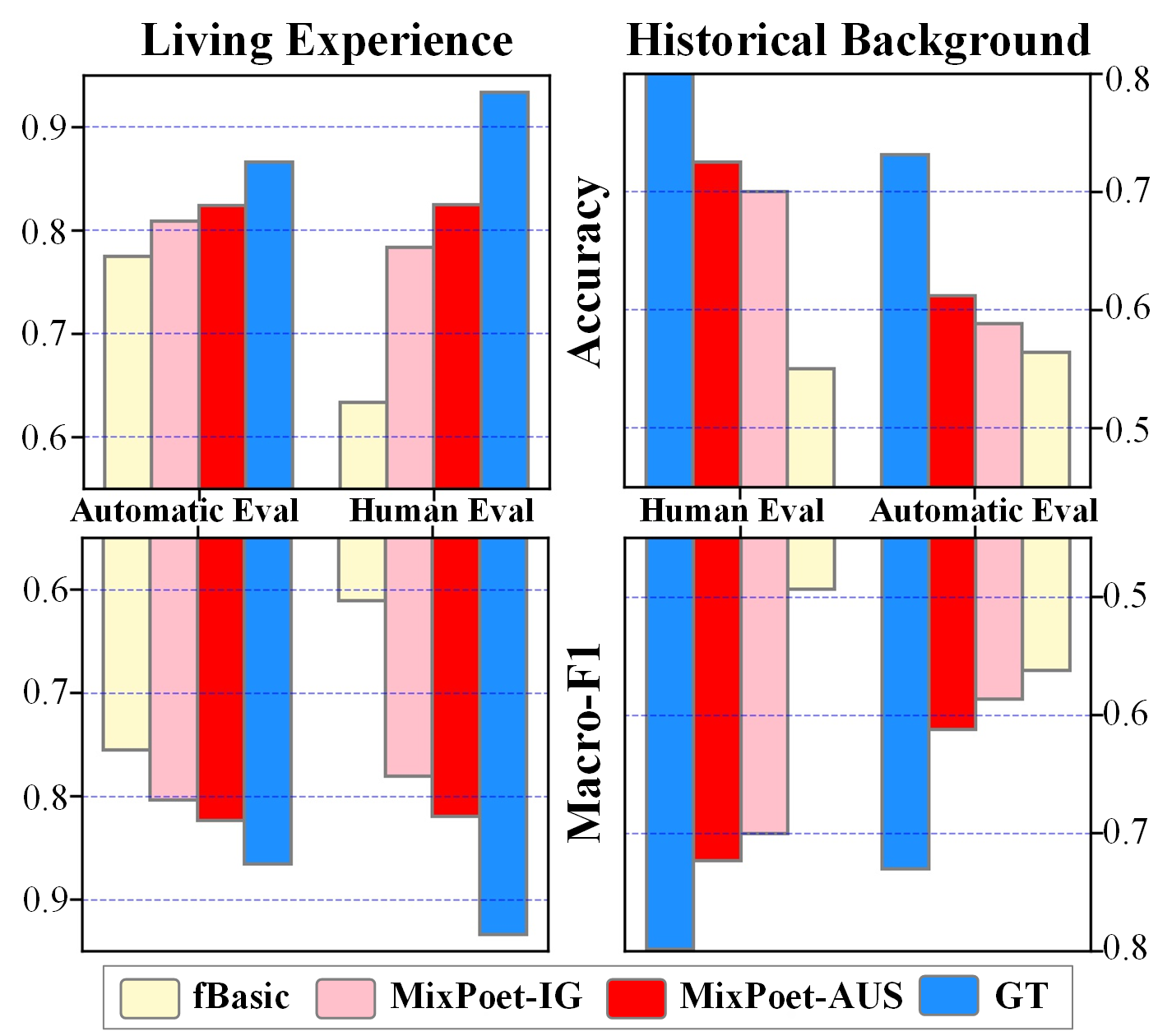}
\caption{Factor control results. We show accuracy and Macro-F1 under both automatic and human evaluations.}
\label{fig_accuracy}
\end{figure}
We set the sizes of hidden state, context vector, latent variable, word embedding and factor embedding to 512, 512, 256, 256 and 64 respectively. The activation function is leaky ReLU for the discriminator and prior networks and is tanh for others. $d \! = \! 3$ in Eq.(4); $\alpha \! = \! \beta=1$ in Eq.(\ref{eq3}). Adam~\cite{Kingmaadam:15} with mini-batches (batch size=128) is used for optimization. To avoid overfitting, we also adopt dropout and $l_2$ norm regularization. For MixPoet-AUS, we update the discriminator five times per update of other parts. We first pre-train our model using both CQC and CQCF, and then fine-tune it with only CQCF. In testing, we adopt beam search (beam size=20) and apply explicit constraints to the search process to ensure that generated poems can meet the requirements of rhyme and rhythm. For fairness, all baselines share the same configuration.
\subsection{Baselines for Comparisons}
\begin{table*}[htp]
\center
\small
\begin{tabular}{c|c|l|l|l|l|l|l}
\Xhline{1.2pt}
Sets & Models  & Fluency  & Coherence  & Meaning & Aesthetics & Relevance & Overall Quality\\
\Xhline{1.2pt}
\multirow{6}{*}{Set 1} & Basic & 3.00 & 2.54 & 2.30 & 2.71 & 2.54 & 2.35 \\
& USPG & 3.09 & 2.65 & 2.61 & 2.98 & 2.73 & 2.63  \\
& CVAE & 3.34 & 2.78 & 2.64 & 3.13 & 2.70 & 2.81 \\
& MRL & 3.91  & 3.66 & 3.36  & 3.73 & 3.19  & 3.55 \\ 
& MixPoet & \bf 4.18$^{**}$  & \bf 4.10$^{**}$ & \bf 3.75$^{**}$  & \bf 4.10$^{**}$ & \bf 3.39 & \bf 3.98$^{**}$ \\ 
\cline{2-8}
& GT &  4.25  &  4.36$^{+}$  & 4.19$^{++}$  & 4.20 & 3.99$^{++}$ & 4.25$^{+}$ \\
\Xhline{1.2pt}
\multirow{2}{*}{Set 2} & fBasic & 3.26 & 3.28  & 2.75  & 3.25 & 2.36 & 2.96 \\
& MixPoet & \bf 4.08$^{**}$ & \bf 4.28$^{**}$ & \bf 3.85$^{**}$ & \bf 4.12$^{**}$ & \bf 2.92$^{**}$ & \bf 3.96$^{**}$ \\
\Xhline{1.2pt}
\end{tabular}
\caption{Human evaluation results of quality. Set 1: poems generated without manually specified mixtures. USPG and MixPoet infer appropriate labels by themselves in terms of different keywords; Set 2: the ones generated with the six mixtures (we present the average scores of all mixtures). Diacritic ** ($p < 0.01$) indicates MixPoet significantly outperforms baseline models; + ($p < 0.05$) and ++ ($p < 0.01$) indicate GT significantly outperforms all models. The Quadratic Weighted Kappa of human annotations is 0.67, which indicates acceptable inter-annotator agreement.}
\label{table:poem_quality}
\end{table*}
We compare the following baselines\footnote{Since our model supports a single keyword, for fairness, we remove the keyword extension module of CVAE and fBasic.}:

\textbf{GT}: ground truth, \textit{i.e.} human-created poems. \textbf{Basic}: the generator introduced in Sec.~\ref{subsec_basic}. \textbf{CVAE} \cite{Yangvaepoetry:18}: a conditional VAE with a hybrid decoder for poetry generation. \textbf{USPG} \cite{Yang:18a}: an unsupervised stylistic poetry generator which supports ten styles and can automatically infer an appropriate style by the input. \textbf{MRL} \cite{Yimrl:18}: a reinforcement learning model which achieves the so-far best inter-topic diversity. \textbf{fBasic}~\cite{Weipoetstyle:18}: a supervised stylistic poetry generator. We also pre-train fBaisc with both CQC and CQCF, and then fine-tune it with CQCF. fBasic takes a straightforward structure, but it represents the typical supervised paradigm of style control.
\subsection{Diversity Evaluation}
As in \cite{Yimrl:18}, we use Jaccard Similarity (JS) to evaluate diversity automatically. For \emph{inter-topic} diversity, we generate 4,500 poems with different keywords but not any manually specified style/mixture, and then calculate JS of them. For \emph{intra-topic} diversity, we calculate JS of poems generated with the same keyword but different specified styles. Besides, to prevent these models cheating by producing ill-formed content, we test the Language Model Score (LMS)~\cite{Yimrl:18} of generated poems. Higher LMS indicates moderate fluency closer to human-authored poetry.

Table \ref{table:diversity} shows that on inter-topic diversity, our model outperforms most baselines and gets very close to MRL. Though with distinct keywords as input, most models tend to generate repetitive phrases (see Figure \ref{fig_case}) which inevitably worsen diversity. MixPoet and USPG incorporate diverse styles to further differentiate generated poems. However, the unsupervised design of USPG results in indistinguishable and uninterpretable learned styles which have no explicit semantic meaning and are too similar. Consequently, even with fewer styles (3*2 vs. 10), MixPoet still surpasses USPG.

MRL obtains the best inter-topic diversity by penalizing high-frequency words but fails to achieve intra-topic diversity. If without extra post-processing, MRL (and Basic) can only generate the same poem by a given keyword (equivalent to intra-JS=100\%). CVAE could produce somewhat different poems by utilizing different samples of $z$, but these poems heavily overlap with each other (intra-JS=38.2\%). We can also see MixPoet-AUS gets better diversity than MixPoet-IG, as the former can learn more discriminable latent mixtures, we will analyse more in Sec.~\ref{subsec_analyse}.
\subsection{Factor Control Evaluation}
\label{subsec_control}
Compared to USPG and MRL, our model attributes diversity to the differences of various styles and interprets each style as a mixture of factor properties. Therefore, we also test if the generated poems are consistent with given factor classes.

For automatic evaluation, we generate 4,000 poems with each mixture and different keywords. Then we use a strong semi-supervised classifier~\cite{Miyatoclassify:16}, which achieves 0.87 and 0.74 F1 values for the two factors respectively, to measure the accuracy. For human evaluation, we generate 20 poems with each mixture (20*6 in total) and invite experts to identify the classes.

As shown in Figure~\ref{fig_accuracy}, fBasic, the typical supervised method, performs the worst due to the quite limited and sparse labelled data. Benefiting from the semi-supervised structure, our model gets noticeable improvement. More than 80\% and 60\% of the generated poems meet specified classes of the two factors, respectively. Such results manifest that, to some extent, a poem generated by MixPoet can simultaneously express the properties of multiple factors.
\subsection{Poetry Quality Evaluation}
\label{subsec_quality}
Since automatic metrics (\textit{e.g.}, perplexity and BLEU) deviate from the human evaluation manner~\cite{Yimrl:18}, we directly adopt human evaluation to assess quality. Following \cite{Yanpolish:16,Zhangmemory:17,Yimrl:18}, we consider: \textbf{Fluency} (is the generated poem well-formed?), \textbf{Context Coherence} (is the poem as a whole thematically and logically structured?), \textbf{Meaningfulness} (does the poem convey certain messages?), \textbf{Aesthetics} (does the poem have some poetic  and artistic beauties?), \textbf{Topic Relevance} (is the poem consistent with the given topic word?) \textbf{Overall Quality} (the general impression on the poem). Each of the six criteria is scored on a 5-point scale ranging from 1 to 5.

We use MixPoet-AUS, which achieves better results in the above assessments, for human quality evaluation and subsequent analyses and refer to it as MixPoet. Then for each model, we generate 40 poems with different randomly-selected keywords. For GT, we choose poems containing corresponding keywords. Therefore, we get 240 (40*6) poems in total. Then we invite ten experts to evaluate in a blind review manner. Each poem is randomly assigned to two experts, and we average the two scores to mitigate personal biases. We refer to \cite{Zhangmemory:17,Yimrl:18} for more details of the evaluation protocol.

As shown in Table \ref{table:poem_quality} (Set 1), MixPoet gets notable improvement compared to other models. USPG is only better than Basic since it adopts a quite simple structure, even without any design for Coherence, which severely limits its performance. CVAE heavily relies on the support of multiple keywords. With a single keyword, it fails to produce meaningful contents, while our model can enrich semantic meanings by the mixed latent space. Despite obtaining the best inter-topic diversity, MRL may lose control of generated contents. Merely increasing TF-IDF could incur unexpected words digressing from topics and thus hurt quality.

Generally, we can find models achieving better diversity (MixPoet and MRL) outperform the others by a large margin since repetitive and generic words can damage poetic images and aesthetic features of generated poems, indicating that diversity also plays a crucial role in promoting quality.

It is noteworthy that generated poems take the risk of straying the given topic when constrained on one single style, because not all topics are compatible with every style. Therefore we also assess poems generated in Sec. \ref{subsec_control}. It can be seen from Table \ref{table:poem_quality} (Set 2) that both fBasic and MixPoet performs somewhat worse on Relevance. Nonetheless, our model still gets acceptable results, since it utilizes the mixed latent space to capture more generalized properties of both factors and keywords, beyond simple labels.
\subsection{Further Analyses}
\label{subsec_analyse}
\begin{figure*}[ht]
\centering
\includegraphics[scale=0.34]{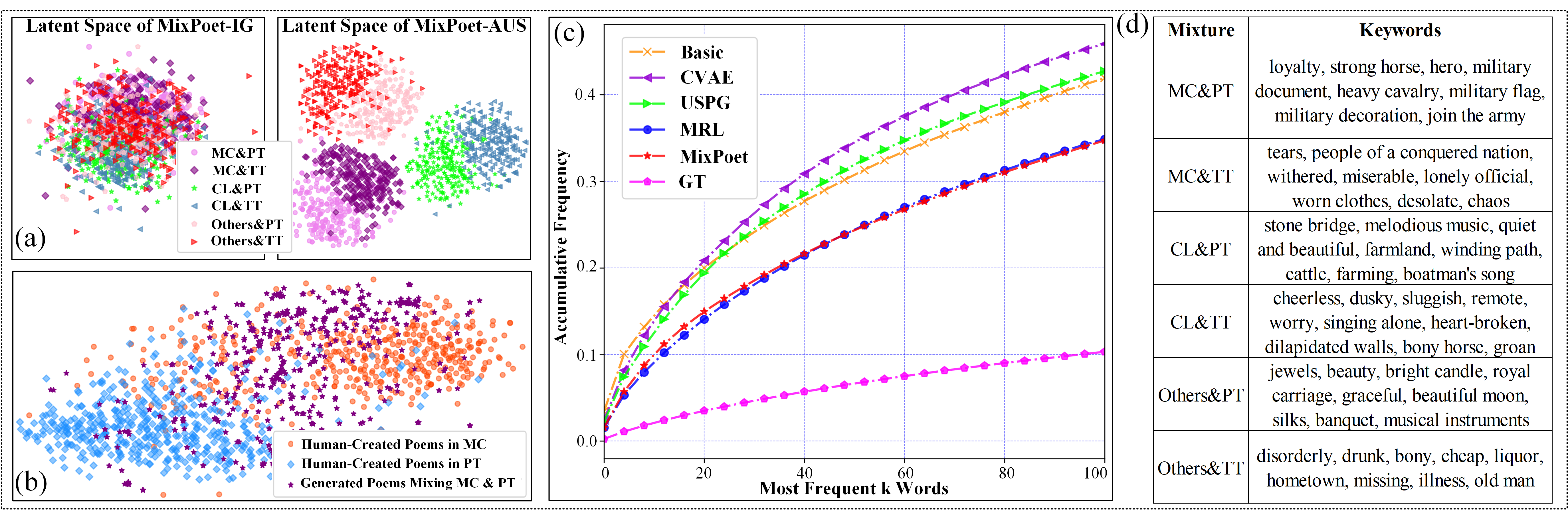}
\caption{(a) Visualization of samples of $z$ conditioned on the keyword `spring wind' and different mixtures. (b) Visualization of human-created and MixPoet-generated poems. (c) Accumulative frequency of the most frequent k words in generated poems. (d) Keywords with the highest prediction probability for each mixture.}
\label{fig_analyse}
\end{figure*}
In Figure \ref{fig_analyse} (a), we visualize points sampled from the prior distributions conditioned on the six mixtures. We can find MixPoet-AUS learns more discriminable latent representations, but Mixpoet-IG fails to distinguish different mixtures.

In Figure \ref{fig_analyse} (b), we vectorize poems by a neural language model and visualize them. We can see poems generated by our model, which mixes two factors (MC\&PT), covers and bridges the two regions of human-authored poems, which indicates that our model successfully achieves the mixture not only on the latent space but also on generated poems.

From Figure \ref{fig_analyse} (c), we can observe that for Basic, CVAE and USPG, a few most frequent words account for a large proportion of generated contents, which leads to quite poor diversity. For instance, the most frequent five words cover over 10\% of all contents generated by Basic. By contrast, our model alleviates this problem and gets a better balance in word distribution.

Figure \ref{fig_analyse} (d) demonstrates the effectiveness of the classifiers involved in our model. We can also find the styles of mixed factors are expressed through concrete contents (\textit{e.g.}, the use of images), which could support our claim that style is coupled with semantics.
\begin{figure}
\centering
\includegraphics[scale=0.35]{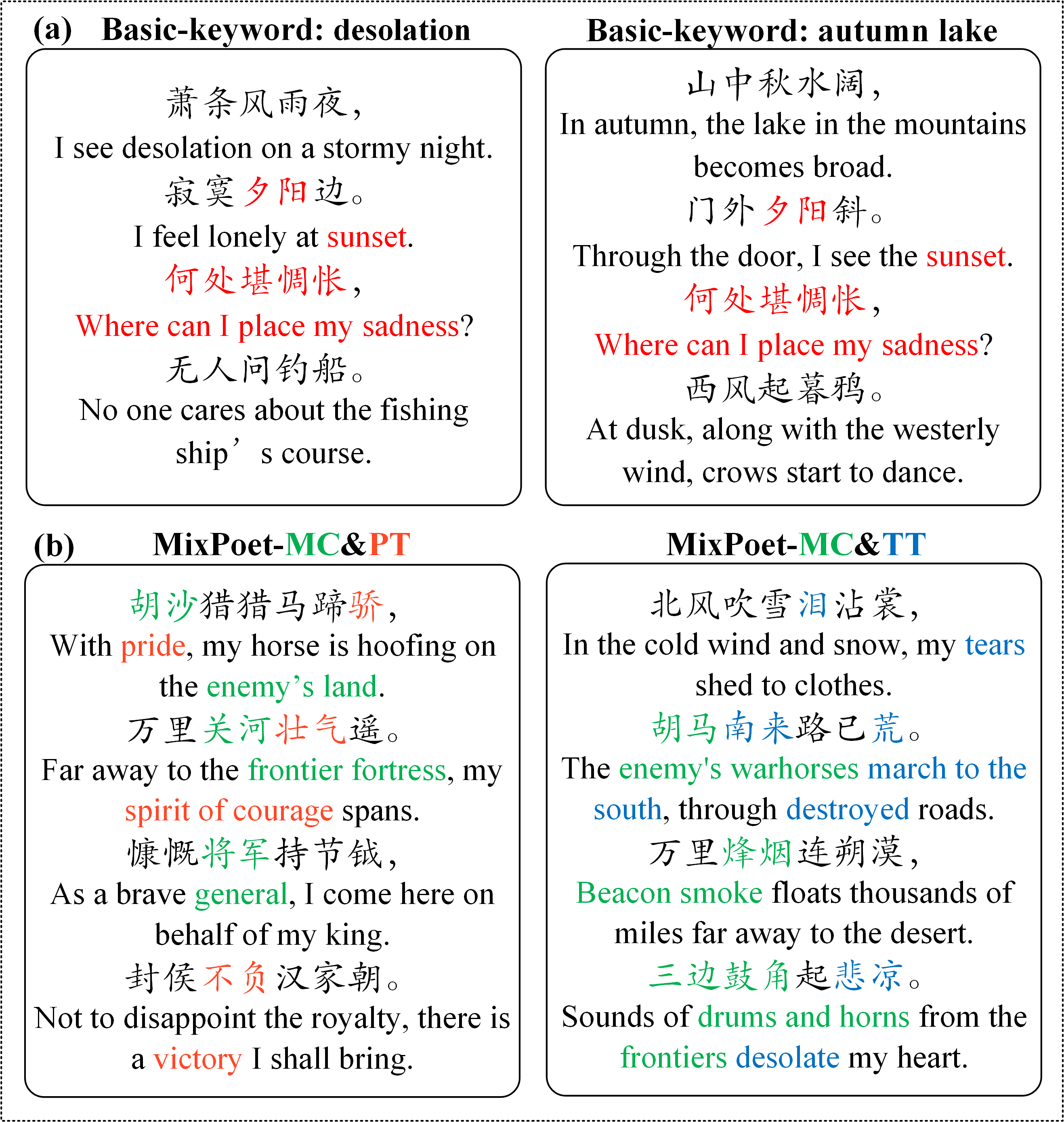}
\caption{(a) Two poems generated by Basic using different keywords. Repetitive phrases are marked in red.  (b) Using the same keyword, two poems generated by MixPoet with different mixtures. Phrases meeting different factor classes are marked in corresponding colors.}
\label{fig_case}
\end{figure}

As shown in Figure \ref{fig_case} (a), with two distinct keywords, Basic generates some repetitive words and even identical whole lines, causing poor diversity. By contrast, in Figure \ref{fig_case} (b), the poem generated by Mixpoet with MC\&PT expresses great heroism and confidence in victory, while the other generated with MC\&TT describes a scene of desolation and shows some loneliness. Besides, in ancient China, some weak dynasties were invaded by northern countries and thus moved their capitals to the south, with which many refugees also fled to the south. MixPoet may capture such events that are widely described by ancient poets and then generates ``enemy's warhorses march to the south'' in the second poem (line 2). Though generated using the same keyword, these two poems present further diversity of thoughts and feelings.
\section{Conclusion and Future Work}
In this work, inspired by related literature theories, we propose \emph{MixPoet}\footnote{MixPoet will be incorporated into \emph{Jiuge}, the THUNLP online poetry generation system (https://jiuge.thunlp.cn).} to address the problem of poor diversity in poetry generation. Based on a semi-supervised VAE, our model disentangles the latent space into different subspaces with each conditioned on one factor which influences human poetry composition. In this way, the generated poems can simultaneously express mixed properties of multiple factors to some degree. By varying the mixture for the same keyword or inferring appropriate factor classes with different keywords, our model differentiates generated poems and hence promotes intra-/inter-topic diversity and quality against three state-of-the-art models.

In the future, we will endeavor to incorporate more factors, such as love experience, school of literary and gender, with finer-granularity discretization. We will also consider modeling the dependence of influence factors, since some factors may be correlative with each other, \textit{e.g.}, gender and living experience, and then apply our model to other kinds of text like story and essay.
\section*{Acknowledgments}
This research is supported by the Major Program of the National Social Science Fund of China (Project No.18ZDA238).

\bibliographystyle{aaai}
\bibliography{aaai20}
\end{document}